\documentclass[runningheads]{llncs}
\usepackage[T1]{fontenc}
\usepackage{graphicx}
\usepackage{multirow}
\usepackage{marvosym}
\usepackage{orcidlink}
\usepackage{caption}
\usepackage{amsmath}
\usepackage{comment}
\usepackage{amssymb}
\usepackage{booktabs} 
\usepackage{float}
\usepackage{tikz}
\usepackage{pgfplots}
\usepackage{microtype}
\usepackage{todonotes}

\pgfplotsset{compat=1.18}

\begin{document}
\title{Task-Oriented Rank Adaptation for\\ Continual Learning in Text Classification}
\titlerunning{Task-Oriented Rank Adaptation}
%
%
\authorrunning{R. Sánchez et al.}
%
%

\author{Rey Sanchez Lopez~\inst{1}
\and
Eduardo Morales Manzanares\inst{1}
\and
Hugo Jair Escalante\inst{1,2}}

%
%
\institute{$^1$INAOE, Puebla, Mexico\\  $^2$The University of Texas at El Paso, El Paso, TX, USA\\ \email{rey.rsl@inaoep.mx}
}

\maketitle              

\begin{abstract}

Continual learning (CL) in text classification faces two critical challenges: catastrophic forgetting and negative transfer across sequential tasks. Parameter-Efficient Fine-Tuning (PEFT) methods such as LoRA enable efficient adaptation by learning low-rank updates of the model parameters. However, these compact representations are normally trained in isolation, limiting their reuse across related tasks. We introduce Task-Oriented Rank Adaptation (TORA), a geometric routing framework that leverages the low-rank structure of LoRA adapters to decide whether to transfer knowledge from the most compatible expert (\textit{Boosting}) or isolate the new task (\textit{Shielding}) based on structural similarity. Evaluated across 15 diverse text classification benchmarks, TORA consistently avoids harmful routing decisions: compatible tasks exceed their isolated performance while reducing training time, and structurally distant tasks are protected from interference with no loss in accuracy. With a single geometric threshold and no reliance on task identities or predefined sequences, TORA provides a simple and effective approach for dynamic adapter routing in sequential text classification systems.

\end{abstract}

\section{Introduction}

Transformer architectures have established the fundamental baseline for Text Classification applications~\cite{vaswani2017attention}. However, extending these architectures to operate in sequential and dynamic environments introduces the Continual Learning (CL)  problem, a machine learning paradigm that involves training a model on a sequential stream of tasks~\cite{de2021continual}. The primary objective is to adapt to new domains while maintaining high performance on previously seen tasks, even when access to historical training data is no longer available.

When a neural network is trained incrementally across multiple domains, it is highly susceptible to catastrophic forgetting \cite{mccloskey1989catastrophic}, a phenomenon where learning new representations overwrites and degrades previously learned parameters. Furthermore, adapting the model to new data distributions often leads to negative transfer~\cite{wang2019characterizing}, where the interference between unrelated tasks deteriorates the overall accuracy and stability of the system.

To mitigate catastrophic forgetting, recent literature has heavily adopted Parameter-Efficient Fine-Tuning (PEFT)  mechanisms~\cite{ding2023parameter}. Specifically, Low-Rank Adaptation (LoRA)  encapsulates task-specific learning into small, low-dimensional weight matrices~\cite{hu2021lora}. By keeping the original pre-trained model frozen, LoRA perfectly protects the base knowledge from being overwritten. 

However, this architectural separation also becomes a major limitation in continual learning scenarios, because these standard PEFT modules operate in strict isolation, the learned representations remain stored within their specific adapters. This strict separation prevents positive transfer between tasks with similar data distributions~\cite{caruana1997multitask}. Consequently, models are forced to learn similar concepts from \textit{Scratch} for every new task, ignoring the computational advantages of knowledge reuse and unnecessarily increasing the number of parameters.

We introduce Task-Oriented Rank Adaptation (TORA), a dynamic routing framework that overcomes the limitations of static adapter isolation in continual learning. In TORA, each previously learned LoRA adapter is treated as an expert, and the shared SVD reference space is constructed from the collection of all learned experts. Incoming tasks are projected into this common geometric space, where structural affinity is estimated through cosine similarity~\cite{mikolov2013distributed} between the new task and the stored experts. When affinity is detected, the router transfers knowledge from the most compatible expert to accelerate convergence (\textit{Boosting}); otherwise, it enforces parameter isolation to prevent negative interference (\textit{Shielding})~\cite{pruksachatkun2020intermediate}. We validate this framework through a geometric analysis of 15 text classification benchmarks and a Monte Carlo simulation, demonstrating empirical advantages over static baselines in both accuracy and parameter efficiency.

The main contributions of this work are summarized as follows:

\begin{itemize}
    \item We propose \textbf{Task-Oriented Rank Adaptation (TORA)}, a geometric routing framework that estimates task affinity directly from LoRA adaptation weights in a shared SVD reference space, eliminating the need for semantic routing or trainable gating networks.

    \item We introduce a compact geometric task fingerprint that represents each LoRA adapter as a low-dimensional structural signature. These fingerprints enable efficient similarity estimation while avoiding computationally expensive manifold-based comparisons, making dynamic routing practical for continual learning.

    \item We validate the proposed framework on 15 text classification benchmarks and Monte Carlo simulations, demonstrating improved routing reliability, positive transfer, and computational savings over isolated training.
\end{itemize}

\section{Related Work}

In text classification, CL addresses the challenge of training models on sequential data streams without suffering from catastrophic forgetting. Early measures primarily focused on regularization, which penalizes changes to essential weights \cite{kirkpatrick2017overcoming}, and rehearsal mechanisms, which store \cite{chaudhry2019tiny} or generate \cite{sun2020lamal} examples of past domains. Although these methods prove effective, they impose limitations in memory and computational cost. Conversely, recent paradigms highlight a positive aspect of sequential learning: accumulating knowledge across multiple domains can actively enhance a model's few-shot adaptation capabilities \cite{ke-etal-2022-continual}. Building upon these insights, the proposed TORA framework transitions from expensive data-replay and complex regularization strategies to a purely parameter-efficient geometric routing. By evaluating structural task affinities, TORA dynamically transfers past knowledge to accelerate convergence on related domains, while strictly isolating incompatible parameters to prevent interference.

To reduce the cost of adapting large language models, PEFT methods such as LoRA have become standard. LoRA freezes the base model and learns low-rank updates, allowing multiple tasks to be trained one after another in the same model. This makes training more efficient and scalable. However, it still does not define how knowledge should be shared or separated between tasks.

Recent research has proposed dynamic architectural approaches to overcome static isolation. Methods such as orthogonal adaptation \cite{wang-etal-2023-orthogonal} and multi-task dynamic routing \cite{wang2022dualprompt} attempt to project new knowledge into distinct subspaces or learn unified representations. While these approaches improve parameter distribution, they often rely on trainable routing mechanisms or computationally expensive similarity estimation, and still lack a simple criterion to evaluate the structural compatibility between sequential tasks before initialization. This reveals a gap for a lightweight routing strategy that directly exploits the geometry of adapter parameters. TORA addresses this gap by representing each task as a compact geometric fingerprint in a shared SVD reference space, enabling deterministic structural similarity estimation without semantic representations, trainable routing modules, or computationally expensive manifold operations.

\section{The TORA Framework}

\begin{figure}[htbp]
    \centering
    \includegraphics[width=1\textwidth]{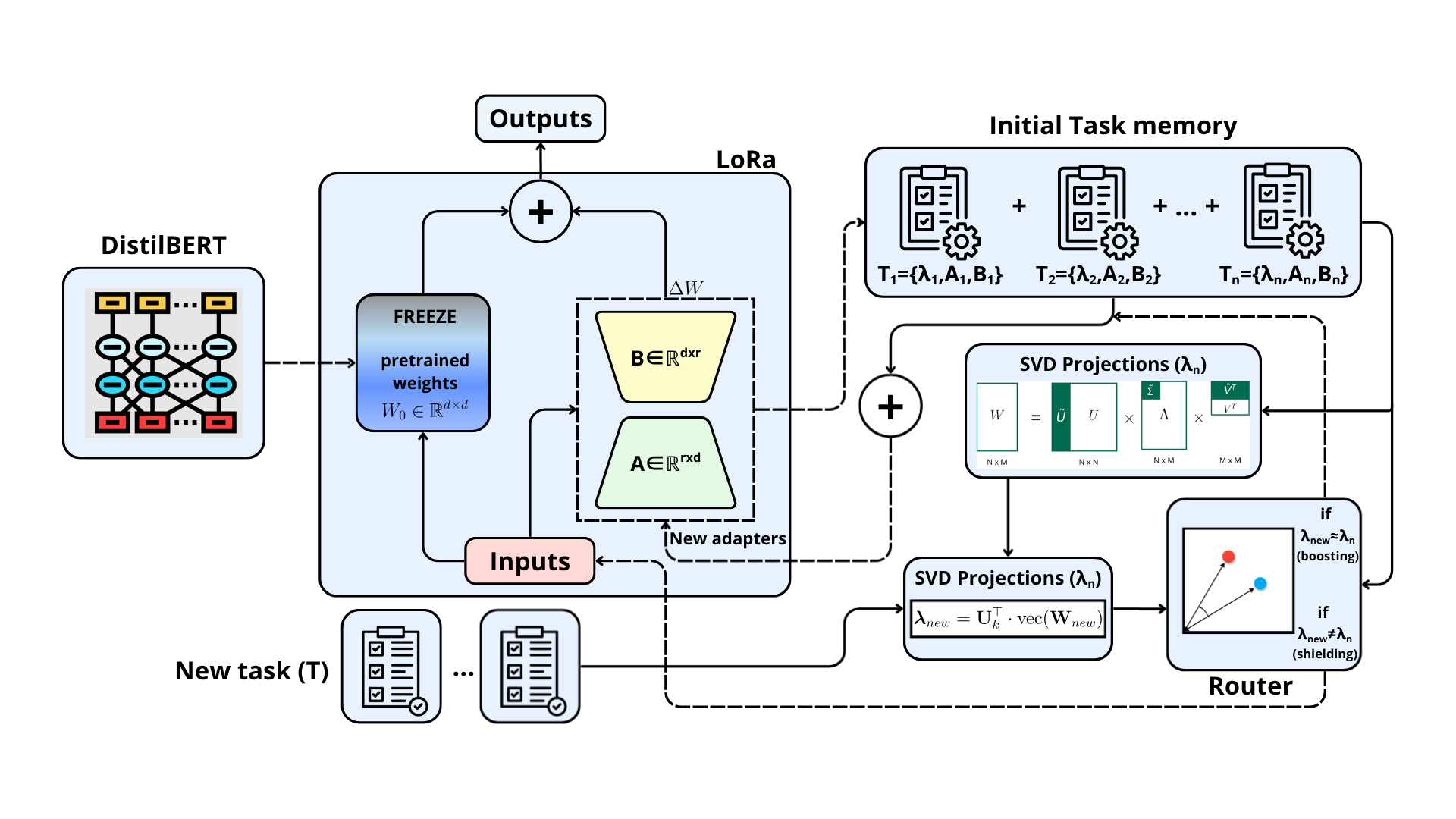}
    \caption{Architectural diagram of the Task-Oriented Rank Adaptation (TORA) framework. The diagram illustrates the routing process, structural affinity evaluation and the memory of task experts.}
    \label{fig:tora_diagram}
\end{figure}

TORA evaluates the geometric compatibility between sequential domains before parameter initialization, replacing the default assumption of task isolation in PEFT methods. An \textit{expert} is defined as a task-specific LoRA adapter trained on a given domain and stored in a memory pool, encapsulating its optimized weights without interfering with previously learned distributions. Figure~\ref{fig:tora_diagram} illustrates the high-level operation of the framework.

Specifically, the continual learning cycle works as follows: when a new task arrives, it is trained for one epoch to get a partial adapter. The SVD of its weights is then computed to extract its geometric fingerprint ($\boldsymbol{\phi}_{new}$). This 
representation is compared against the reference space built from the previously stored experts. If the structural similarity is above the angular threshold, the router triggers a transfer protocol, loading the weights of the most compatible expert as a starting point for the new task. If not, the shielding protocol is activated and new adaptation  matrices ($A$ and $B$) are trained from scratch. Once training is complete, the resulting adapter is saved to the memory pool, updating the reference space for any future tasks.


The framework builds upon the mathematical foundation of LoRA. During standard training, LoRA freezes a pre-trained weight matrix $W_0 \in \mathbb{R}^{d \times d}$ and constrains its update by representing it with a low-rank decomposition $\Delta W = BA$. The matrices are defined as $B \in \mathbb{R}^{d \times r}$ and $A \in \mathbb{R}^{r \times d}$, where the rank $r \ll d$. The forward pass is thus modified to:
$$W = W_0 + \Delta W = W_0 + BA$$

Within this formulation, matrix $A$ acts as a feature extractor that maps the input down to a lower-dimensional, information-dense subspace. Matrix $B$ subsequently projects these representations back to the original dimension. By capturing task-specific knowledge in these low-rank matrices, the model significantly reduces the number of trainable parameters.

To enable dynamic routing, TORA evaluates the structural affinity of these learned projections. During the training phase of a new task, TORA applies SVD to the existing adapter weight matrices to isolate their core semantic directions. For a given adaptation matrix $\Delta W$, the decomposition is defined as:
$$\Delta W = U \Sigma V^T$$
where the orthogonal matrix $U$ contains the left singular vectors that define the geometric rotation in the output space, the diagonal matrix $\Sigma$ contains the singular values that quantify the magnitude of variance along each principal direction, and the matrix $V$ contains the principal components that define the spatial orientation of the learned task.


\subsection{Reference Space Construction}\label{subsec:Reference Space Construction}

To determine geometric compatibility, the system first builds a shared geometric space from the $N{-}1$ previously learned LoRA adapters, since the incoming ($N$-th) task is the one currently being evaluated and has not yet become an expert. Each expert denotes the complete set of LoRA adaptation parameters associated with a previously learned task. For each layer $l$ and attention projection $p$, the corresponding weight tensors are flattened into vectors and stacked into a variability matrix.

In simple terms, this matrix acts as a consolidated memory bank that aggregates the parametric signatures of all previously learned domains. By grouping these isolated expert vectors together, the resulting matrix captures the complete spatial variance of the model's history. This representation is highly effective because it allows the subsequent SVD to extract the shared principal components creating a unified reference space to evaluate any new incoming task:

\small{
\begin{equation}
    \mathbf{M}^{(l,p)} = \left[ \text{vec}\left(\mathbf{A}_{1}^{(l,p)}\right) \;\Big|\;
    \text{vec}\left(\mathbf{B}_{1}^{(l,p)}\right) \;\Big|\; \cdots \;\Big|\; \text{vec}\left(\mathbf{A}_{N-1}^{(l,p)}\right) \;\Big|\; \text{vec}\left(\mathbf{B}_{N-1}^{(l,p)}\right) \right] \in \mathbb{R}^{D \times (N-1)}
    \label{eq:matriz_variabilidad}
\end{equation}}

\noindent where $D = d \times r$ and $\text{vec}(\cdot)$ flattens a matrix 
into a column vector. After row-wise centering, SVD is applied:

\begin{equation}
    \tilde{\mathbf{M}}^{(l,p)} = \mathbf{U}^{(l,p)}\, \mathbf{S}^{(l,p)}\, 
    {\mathbf{V}^{(l,p)}}^\top
    \label{eq:svd}
\end{equation}

\noindent \normalsize{The left singular vectors $\mathbf{U}^{(l,p)}$ form the orthonormal 
basis of the reference space \cite{gur2018gradient}. Only the first $k$ components are retained, 
where $k$ is the minimum number of directions that explain 60\% of the 
total variance:}

\begin{equation}
    k = \min\!\left\{ j \;\middle|\; 
    \frac{\sum_{i=1}^{j} s_i^2}{\sum_{i=1}^{N-1} s_i^2} \geq 0.60 \right\}
    \label{eq:k}
\end{equation}

\noindent In the context of a continuous stream of $N$ domains, this formulation establishes an iterative routing cycle. At any given step, the memory pool maintains $N-1$ established expert representations to evaluate a single incoming task. Following adaptation, the newly processed task is integrated into the expert memory, dynamically updating the reference space for the subsequent iteration.

\vspace{-0.5cm}

\vspace{-0.9cm}

\subsection{Task Fingerprint Generation}

To characterize a task in the reference space, its LoRA weights are projected onto the $k$ selected singular vectors as:

\begin{equation}
\boldsymbol{\lambda}^{(l,p)} = \mathbf{U}_k^{(l,p)\top} \cdot
\text{vec}\!\left(\mathbf{W}^{(l,p)}\right)
\label{eq:loadings}
\end{equation}

\noindent where $\mathbf{W}^{(l,p)}$ denotes the LoRA weight matrix at layer $l$ and projection $p$, $\text{vec}(\cdot)$ vectorizes it, and $\mathbf{U}_k^{(l,p)}$ contains the top-$k$ singular vectors obtained via SVD from historical task representations (Section \ref{subsec:Reference Space Construction}). The resulting $\boldsymbol{\lambda}^{(l,p)} \in \mathbb{R}^k$ encodes the alignment of the task with the principal directions of the reference space.

The task fingerprint is then constructed by concatenating these projections across all adaptation weights. In the case of DistilBERT, which comprises 6 Transformer layers with LoRA modules applied to the (\texttt{q\_lin}, \texttt{k\_lin}, and \texttt{v\_lin}) projections, this results in 18 modules and 36 parameter matrices ($A$ and $B$ per module).

By projecting each of these 36 matrices onto its $k=4$ most relevant singular vectors (the eigenvectors that satisfy the variance threshold), the model is compressed from its high-dimensional parameter space into an exact coordinate:

\begin{equation}
    \boldsymbol{\phi} = \left[ \boldsymbol{\lambda}^{(1,\text{q})}_A \;\|\; 
    \boldsymbol{\lambda}^{(1,\text{q})}_B \;\|\; \cdots \;\|\; 
    \boldsymbol{\lambda}^{(6,\text{v})}_B \right] \in \mathbb{R}^{144}
    \label{eq:fingerprint}
\end{equation}

\noindent This 144-dimensional vector ($36 \text{ sub-matrices} \times 4 \text{ principal components}$) encodes the absolute position of the task in the shared SVD space, and serves as the compact geometric signature used by the router to compare structural affinities across datasets.


\subsection{Routing Mechanism with Shielding and Boosting}

The routing mechanism operates entirely \textit{a priori}: the transfer decision is made 
before completing the full training of the incoming task, incurring only the cost of a 
single training epoch (approximately 10--12 seconds). Let $\mathcal{E} = \{e_1, e_2, 
\ldots, e_{N-1}\}$ denote the set of $N-1$ expert adapters stored in memory. The 
procedure is as follows:

\begin{enumerate}

    \item \textbf{Initial training.} The incoming task $\mathcal{T}_\text{new}$ is trained 
    for 1 epoch, yielding a partial adapter with weights 
    $\{\mathbf{A}_\text{new}^{(l,p)},\, \mathbf{B}_\text{new}^{(l,p)}\}$.

    \item \textbf{SVD space construction.} The reference space is built from the $N-1$ 
    stored experts following the procedure described in Section~\ref{subsec:Reference Space Construction}.

    \item \textbf{Fingerprint computation.} Let $\boldsymbol{\phi}_\text{new}$ represent the extracted principal component vector of the incoming task, and $\boldsymbol{\phi}_i$ represent the vectors of the previously learned experts. To obtain these representations, the partial adapter of the incoming task (trained for 1 epoch) and the adapters of all experts (trained for 20 epochs) are projected onto the shared SVD space, yielding the final dimensional fingerprints $\boldsymbol{\phi}_\text{new}, \boldsymbol{\phi}_i \in \mathbb{R}^{144}$ for each expert $e_i \in \mathcal{E}$.

    \item \textbf{Similarity ranking.} Subsequently, the framework employs cosine similarity as the parametric distance metric to quantify the geometric overlap between representations. The structural affinity dictates the adaptation strategy through the following protocol: \\
    The cosine similarity and angular distance between 
    the incoming task and each expert are computed as:
    \begin{equation}
        \theta_i = \arccos\!\left(\frac{\boldsymbol{\phi}_\text{new} \cdot 
        \boldsymbol{\phi}_i}{\|\boldsymbol{\phi}_\text{new}\|\,\|\boldsymbol{\phi}_i\|}\right) 
        \cdot \frac{180^{\circ}}{\pi}, \quad i = 1, \ldots, N-1
        \label{eq:angulo}
    \end{equation}
    The experts are then sorted in ascending order of $\theta_i$, defining a ranking 
    $\sigma$ such that:
    \begin{equation}
        \theta_{\sigma(1)} \leq \theta_{\sigma(2)} \leq \cdots \leq \theta_{\sigma(N-1)}
        \label{eq:ranking}
    \end{equation}
    where $\sigma(k)$ denotes the index of the $k$-th most similar expert. In particular, 
    $\sigma(1)$ identifies the \textbf{Top-1 expert}: the stored adapter whose weight 
    space is geometrically closest to the incoming task.

    \item \textbf{Routing decision.} 
    
    The mean angular distance over the Top-3 most similar experts is used as the decision router tool, derived in the Section~\ref{angular_mean}  and is calculated as:
    \begin{equation}
        \bar{\theta}_3 = \frac{1}{3}\sum_{k=1}^{3} \theta_{\sigma(k)}
        \label{eq:promedio}
    \end{equation}
    This average characterizes the \textit{geometric neighborhood} of the incoming task. 
    The routing decision is then:
    \begin{equation}
        \mathbf{W}_\text{init} =
        \begin{cases}
            \mathbf{W}_{\text{exp}(\sigma(1))} & \text{if } \bar{\theta}_3 < 84.5^{\circ}
            \text{ \ (TORA: Top-1 expert transfer)} \\
            \mathbf{W}_0 & \text{if } \bar{\theta}_3 \geq 84.5^{\circ}
            \text{ \ (Shielding: train from scratch)}
        \end{cases}
        \label{eq:decision}
    \end{equation}
    where $\mathbf{W}_{\text{exp}(\sigma(1))}$ are the weights of the Top-1 expert and 
    $\mathbf{W}_0$ denotes randomly initialized LoRA weights.

    \item \textbf{Full training.} The incoming task is trained to completion using 
    $\mathbf{W}_\text{init}$ as initialization. Upon convergence, the resulting adapter 
    is stored in memory as a new expert available for future routing decisions.

\end{enumerate}


\section{Experimental Setup}

To evaluate the proposed framework across diverse linguistic domains, the experimental setup includes 15 text classification datasets. The selected benchmarks cover a wide range of Text Classification tasks to ensure a robust evaluation of the structural affinity metric across geometrically distinct data distributions, as used in previous work~\cite{wang-etal-2023-orthogonal}.

\subsection{Datasets and Benchmarks}

For reproducibility TORA is evaluated on a set of 15 text classification tasks derived from three sources: Yelp Polarity, Amazon Polarity, Yahoo Answers, and AG News datasets \cite{zhang2015character}, DBPedia \cite{lehmann2015dbpedia}, GLUE~\cite{wang2018glue}, and SuperGLUE~\cite{wang2019superglue}. Table~\ref{tab:tareas} summarizes the selected tasks.

\subsection{Implementation Details}

The experiments use DistilBERT~\cite{sanh2019distilbert}, a distilled version of BERT~\cite{devlin-etal-2019-bert} that retains 97\% of its performance with 40\% of the parameters. For the PEFT configuration (Table~\ref{tab:hyperparams}), LoRA modules are injected into the self-attention layers of the Transformer~\cite{lialin2023scaling}.

Following~\cite{hu2021lora}, we set the adaptation rank to $r=64$, as larger ranks yield diminishing returns while lower ranks already capture the relevant adaptation subspace. This choice balances expressivity and efficiency for the 15 heterogeneous NLP tasks considered ($r \ll d=768$).

For DistilBERT ($d=768$), this results in adapter matrices $A \in \mathbb{R}^{64 \times 768}$ and $B \in \mathbb{R}^{768 \times 64}$, reducing the number of trainable parameters per projection from $d^2$ to $2rd$ (i.e., $\sim$6$\times$ reduction). Across all 18 LoRA-injected projections, this yields 1.77M parameters per task, compared to 10.62M for full fine-tuning.

\begin{table}[h]
\centering
\scriptsize
\caption{Evaluation tasks for the TORA framework.}
\label{tab:tareas}
\resizebox{\textwidth}{!}{
\begin{tabular}{llll}
\toprule
\textbf{Task} & \textbf{Benchmark} & \textbf{Type} & \textbf{Description} \\
\midrule
Yelp Polarity   & CL Benchmark & Binary classification & Sentiment in Yelp reviews \\
Amazon Polarity & CL Benchmark & Binary classification & Sentiment in Amazon reviews \\
DBPedia-14      & CL Benchmark & 14-class classification & DBPedia article categories \\
Yahoo Answers   & CL Benchmark & 10-class classification & Topics in Yahoo questions \\
AG News         & CL Benchmark & 4-class classification  & News classification \\
MNLI            & GLUE         & Natural language inference & Entailment/contradiction/neutral \\
QQP             & GLUE         & Paraphrase detection    & Duplicate question detection \\
RTE             & GLUE         & Natural language inference & Binary entailment \\
SST-2           & GLUE         & Binary classification   & Sentiment in movie reviews \\
WiC             & SuperGLUE    & Word sense disambiguation & Word sense in context \\
CB              & SuperGLUE    & Natural language inference & 3-class entailment \\
COPA            & SuperGLUE    & Causal reasoning        & Plausible alternative selection \\
BoolQ           & SuperGLUE    & Binary QA               & Yes/no questions on paragraphs \\
MultiRC         & SuperGLUE    & Multiple-choice QA      & Multi-sentence questions \\
IMDB            & Standalone   & Binary classification   & Sentiment in movie reviews \\
\bottomrule
\end{tabular}
}

\end{table}

\vspace{-46pt}

\begin{table}[h]
\centering
\caption{Experimental hyperparameter configuration.}
\label{tab:hyperparams}
\scriptsize
\begin{tabular}{lc}
\toprule
\textbf{Hyperparameter} & \textbf{Value} \\
\midrule
Base model                  & DistilBERT-base-uncased \\
LoRA rank ($r$)             & 64 \\
LoRA alpha ($\alpha$)       & 16 \\
LoRA dropout                & 0.1 \\
LoRA bias                   & none \\
LoRA target modules         & \texttt{q\_lin, k\_lin, v\_lin} \\
Transformer layers          & 6 \\
Expert training epochs      & 20 \\
Routing epoch               & 1 \\
Learning rate (all tasks)   & $5 \times 10^{-5}$ \\
SVD variance threshold      & 60\% \\
Fingerprint dimensionality  & 144 \\
Angular routing threshold ($\bar{\theta}_3$) & $84.5^\circ$ \\
\bottomrule
\end{tabular}
\end{table}

Within the PEFT configuration (Table~\ref{tab:hyperparams}), we set $\alpha=16$, dropout$=0.1$, and bias=\texttt{none}, following standard LoRA practice~\cite{hu2021lora}. Freezing bias terms ensures that task-specific information is captured exclusively by the low-rank adapters, which is required for our SVD-based analysis. We compare two training settings: \textit{Scratch}, where LoRA adapters are randomly initialized for each task, and \textit{Transfer}, where TORA initializes adapters via the proposed similarity-based routing mechanism.

\section{Evaluation Methodology and Experimental Results}

We evaluate continual learning along two complementary dimensions. First, we analyze temporal dynamics by tracking accuracy across epochs, comparing TORA against a continuous baseline (a single LoRA adapter without parameter isolation). This reveals when positive transfer (Boosting) occurs and confirms effective mitigation of catastrophic forgetting. Second, we assess robustness to task order via a Monte Carlo simulation~\cite{robert2004monte} with $N=100{,}000$ random task sequences, estimating long-term behavior under stochastic arrivals and the probability of transfer or isolation. Based on this, tasks are categorized as structural hubs that enable knowledge sharing or isolated nodes that require shielding.

\subsection{Experimental results}
\label{angular_mean}

Table~\ref{tab:angular_neighbors} shows a geometric structure across the 15 benchmarks. While all tasks remain below $84.5^\circ$ when considering only the nearest neighbor ($\bar{\theta}_1$), progressively larger neighborhoods improve separability. The Top-3 average ($\bar{\theta}_3$) yields the most stable boundary: tasks that benefit from transfer consistently fall below $84.5^\circ$, whereas isolated tasks exceed this threshold. In contrast, smaller neighborhoods are more sensitive to individual neighbor variability, while expanding to larger neighborhoods (e.g., $\bar{\theta}_4$ and $\bar{\theta}_5$) introduces more distant experts into the average, diluting the local geometric structure and leading to misclassifications such as AGNEWS. These results identify $\bar{\theta}_3$ as the most robust routing criterion for the 15-task benchmark considered in this work; consequently, it was selected as the decision metric.

    \begin{table}[h] 
    \centering
    \caption{Cosine similarity (in degrees) between each incoming task and 
    its Top-5 nearest experts, along with cumulative averages 
    $\bar{\theta}_k$ for $k = 1, \ldots, 5$. The routing threshold of $84.5^\circ$ applied to $\bar{\theta}_3$ provides the clearest separation between transfer and shielding decisions.}
    \label{tab:angular_neighbors}
    \scriptsize
    \resizebox{\textwidth}{!}{%
    \begin{tabular}{lccccccccccc}
    \toprule
    \textbf{Task} & $\theta_1$ & $\theta_2$ & $\theta_3$ & $\theta_4$ & 
    $\theta_5$ & $\bar{\theta}_1$ & $\bar{\theta}_2$ & $\bar{\theta}_3$ & 
    $\bar{\theta}_4$ & $\bar{\theta}_5$ & \textbf{Decision} \\
    \midrule
    IMDB    & 75.9348 & 78.2152 & 81.5054 & 81.9472 & 83.3364 & 75.9348 & 77.0750 & 78.5518 & 79.4006 & 80.1878 & Transfer \\
    WIC     & 78.2600 & 80.1432 & 86.1518 & 86.2642 & 86.3813 & 78.2600 & 79.2016 & 81.5183 & 82.7048 & 83.4401 & Transfer \\
    RTE     & 78.6749 & 79.5074 & 87.4172 & 87.7247 & 87.7818 & 78.6749 & 79.0912 & 81.8665 & 83.3311 & 84.2212 & Transfer \\
    COPA    & 79.5401 & 81.8497 & 82.7471 & 82.9056 & 83.0827 & 79.5401 & 80.6949 & 81.3790 & 81.7606 & 82.0250 & Transfer \\
    YELP    & 79.8080 & 84.1105 & 88.0673 & 88.8443 & 91.1074 & 79.8080 & 81.9593 & 83.9953 & 85.2075 & 86.3875 & Transfer \\
    MULTIRC & 80.4721 & 80.7501 & 81.1396 & 82.0910 & 82.0954 & 80.4721 & 80.6111 & 80.7873 & 81.1132 & 81.3096 & Transfer \\
    QQP     & 81.1524 & 81.3621 & 81.6509 & 81.7015 & 81.9840 & 81.1524 & 81.2573 & 81.3885 & 81.4667 & 81.5702 & Transfer \\
    YAHOO   & 81.0267 & 84.4764 & 85.9152 & 86.1951 & 86.6186 & 81.0267 & 82.7516 & 83.8061 & 84.4034 & 84.8464 & Transfer \\
    AMAZON  & 82.0722 & 82.6582 & 84.5824 & 85.4851 & 85.5997 & 82.0722 & 82.3652 & 83.1043 & 83.6995 & 84.0795 & Transfer \\
    BOOLQ   & 82.3213 & 82.3664 & 82.5222 & 82.9515 & 83.2081 & 82.3213 & 82.3439 & 82.4033 & 82.5404 & 82.6739 & Transfer \\
    DBPEDIA & 82.6105 & 82.6738 & 82.7011 & 82.8122 & 83.9575 & 82.6105 & 82.6421 & 82.6618 & 82.6994 & 82.9510 & Transfer \\
    AGNEWS  & 84.0747 & 84.0747 & 85.0874 & 85.1226 & 85.6418 & 84.0747 & 84.0747 & 84.4123 & 84.5898 & 84.8002 & Transfer \\
    \midrule
    SST2    & 83.8398 & 85.1843 & 85.2407 & 85.9529 & 85.9926 & 83.8398 & 84.5121 & \textbf{84.7549} & 85.0544 & 85.2420 & Shielding \\
    CB      & 83.2625 & 85.9329 & 88.2758 & 88.5630 & 89.2346 & 83.2625 & 84.5977 & \textbf{85.8237} & 86.5085 & 87.0538 & Shielding \\
    MNLI    & 84.0934 & 85.4571 & 87.0392 & 89.4239 & 90.3129 & 84.0934 & 84.7753 & \textbf{85.5299} & 86.5034 & 87.2653 & Shielding \\
    \bottomrule
    \end{tabular}%
    }
    \end{table}

\vspace{-1cm}

\subsection{Interference Prevention (Shielding)}

Table~\ref{tab:interference_prevention} shows demonstrates that TORA reliably prevents negative transfer by enforcing isolation when $\bar{\theta}_3 \ge 84.5^\circ$. For CB, MNLI, and SST-2, this avoids degradation caused by incompatible representations. In particular, forcing transfer on MNLI reduces accuracy to 0.628 while doubling training time, whereas Shielding preserves baseline performance (0.652) with efficient 10-epoch training. This confirms that the routing mechanism not only enables transfer when beneficial, but also robustly avoids detrimental interactions.

\subsection{Positive Transfer (Boosting)}
Table~\ref{tab:positive_transfer} shows that TORA effectively exploits geometric affinity ($\bar{\theta}_3 < 84.5^\circ$) to enable positive transfer, surpassing isolated training (\textit{Scratch}). Gains are particularly pronounced in tasks with higher abstraction or limited data. For instance, COPA achieves a $+11.11\%$ improvement while reducing training by 10 epochs when initialized from DBPedia, indicating efficient reuse of geometric aligned representations that accelerates convergence and enhances performance.

\begin{table}[htbp]
\centering

\caption{Interference Prevention: Performance and computational cost of forcing geometric transfer versus applying the Shielding protocol.}
\label{tab:interference_prevention}
\scriptsize 
\resizebox{\textwidth}{!}{%
\begin{tabular}{lccccc}
\toprule
\textbf{Task} & \textbf{Forced Transfer} & \textbf{Scratch (TORA)} & \textbf{Avoided Degradation} & \textbf{Epochs Saved} & \textbf{Routing Decision} \\
\midrule
CB      & 0.820 & \textbf{0.820} & 0.000 & 10 & Isolation (Scratch) \\
MNLI    & 0.628 & \textbf{0.652} & \textbf{+0.024} & 10 & Isolation (Scratch) \\
SST2    & 0.870 & \textbf{0.870} & 0.000 & 10 & Isolation (Scratch) \\
\bottomrule
\end{tabular}%
}

\vspace{0.2cm}

\caption{Positive Transfer: Target tasks that surpassed their asymptotic accuracy limits under the TORA routing mechanism.}
\label{tab:positive_transfer}
\scriptsize 
\resizebox{\textwidth}{!}{%
\begin{tabular}{lccccc}
\toprule
\textbf{Task} & \textbf{Scratch Baseline} & \textbf{TORA Accuracy} & \textbf{Relative Gain} & \textbf{Epochs Saved} & \textbf{Routing Decision} \\
\midrule
COPA    & 0.450 & 0.500 & +11.11\% & 10 & Transfer from DBPEDIA \\
MULTIRC & 0.544 & 0.580 & +6.62\%  & 10 & Transfer from SST2 \\
WIC     & 0.662 & 0.702 & +6.04\%  & 0  & Transfer from DBPEDIA \\
BOOLQ   & 0.632 & 0.656 & +3.80\%  & 10 & Transfer from QQP \\
RTE     & 0.602 & 0.621 & +3.16\%  & 5  & Transfer from YELP \\
AMAZON  & 0.876 & 0.892 & +1.83\%  & 10 & Transfer from AGNEWS \\
QQP     & 0.772 & 0.782 & +1.30\%  & 0  & Transfer from RTE \\
YAHOO   & 0.700 & 0.722 & +1.03\%  & 10 & Transfer from DBPEDIA \\
YELP    & 0.894 & 0.908 & +1.01\%  & 10 & Transfer from CB \\
IMDB    & 0.830 & 0.832 & +0.24\%  & 10 & Transfer from WIC \\
\bottomrule
\end{tabular}%
}

\end{table}

\vspace{-1cm}

\subsection{Temporal Dynamics and Performance}

The geometric classification directly determines convergence dynamics. Table~\ref{tab:positive_transfer} and table~\ref{tab:interference_prevention} shows epoch-wise validation accuracy across the 15 datasets, comparing isolated training against TORA. For geometric hubs (e.g., IMDB, MultiRC, QQP), TORA induces a clear \textit{Boosting} effect: initializing from the nearest expert yields higher early accuracy and faster convergence, indicating effective reuse of aligned representations. For structurally divergent tasks (SST-2, MNLI, CB), TORA activates \textit{Shielding} when $\bar{\theta}_3 \ge 84.5^\circ$, preventing negative transfer. In these cases, isolated initialization preserves baseline performance while avoiding degradation, demonstrating that the routing mechanism both accelerates learning when beneficial and enforces isolation when necessary.

\subsection{Monte Carlo simulations}

To evaluate robustness to task order, we run a Monte Carlo simulation with 100,000 random task sequences (Table~\ref{tab:routing_probabilities_100k}). The results confirm stable routing behavior: dissimilar tasks such as SST-2 and MNLI consistently trigger \textit{Shielding} (99.5\%), while related tasks like COPA achieve successful transfer in 80.8\% of cases. These results indicate that routing decisions are driven by latent-space structure rather than task order, ensuring consistent performance in continual learning settings.

\begin{figure}[h]
\centering

\captionof{table}{Routing probability distribution following Monte Carlo simulation ($N=100,000$).}
\label{tab:routing_probabilities_100k}
\scriptsize
\resizebox{\textwidth}{!}{%
\begin{tabular}{lccc}
\toprule
\textbf{Task (Domain)} & \textbf{Isolation Prob. (Scratch)} & \textbf{Transfer Prob.} & \textbf{Geometric Classification} \\
\midrule
\multicolumn{4}{c}{\textbf{The Isolated Cases (Isolators)}} \\
\midrule
SST2 & \textbf{99.50\%} & 0.50\% & Highly Isolated Node \\
CB & \textbf{99.50\%} & 0.50\% & Highly Isolated Node \\
MNLI & \textbf{99.50\%} & 0.50\% & Highly Isolated Node \\
AGNEWS & \textbf{69.00\%} & 31.00\% & Rigid Knowledge Node \\
\midrule
\multicolumn{4}{c}{\textbf{The Compatible Cases (Hubs)}} \\
\midrule
COPA & 19.20\% & \textbf{80.80\%} & Highly Compatible Node \\
MULTIRC & 19.40\% & \textbf{80.60\%} & Highly Compatible Node \\
BOOLQ & 19.90\% & \textbf{80.10\%} & Central Hub \\
IMDB & 21.80\% & \textbf{78.20\%} & Central Hub \\
\bottomrule
\end{tabular}%
}

\vspace{-1cm}

\end{figure}

\subsection{Continual Learning Results and Dynamic Routing Efficacy}

Since continual learning is highly sensitive to task order, we compare TORA against manually designed sequences derived from analysis: a similar-first (sentiment analysis, followed by topic classification, and finally reasoning tasks) order and its reverse as shown in Table~\ref{tab:sequence_details}. Table~\ref{tab:resultados_cl} shows that task ordering alone has a strong impact on performance, with results ranging from 60.75\% to 66.99\% across our two sequences, and from 64.8\% to 72.3\% across the three handcrafted orders of O-LoRA~\cite{wang-etal-2023-orthogonal}. This confirms that manually defining optimal sequences is not scalable. In contrast, TORA removes this dependency entirely by dynamically deciding when to transfer or isolate knowledge based on geometric affinity, achieving $A_T = 75.56\%$ at best and $A_T = 75.40\%$ on average across all five evaluated sequences, outperforming O-LoRA in every tested order while exhibiting significantly lower variance across sequences (0.33\% (TORA) vs 7.5\% (O-LoRA)).

\section{Discussion}

The results show a clear and consistent pattern across all 15 evaluated tasks. TORA never made a routing decision that hurt performance: tasks that were geometrically close to existing experts benefited from 
knowledge transfer, while structurally distant tasks were protected by the Shielding protocol. This was achieved without any task identity information or manually designed sequences, which suggests that the angular distance in the SVD parameter space is a reliable indicator of task compatibility across diverse text classification domains. The following subsections discuss the theoretical and practical implications of this finding.

\begin{table}[!ht]
\centering
\scriptsize

\caption{Continual learning performance comparison across 15 text classification tasks. We report the Average Final Accuracy ($A_T$) for different sequence configurations.}
\label{tab:resultados_cl}
\begin{tabular}{lcc}
\toprule
\textbf{Method / Configuration} & \textbf{Order} & \textbf{$\boldsymbol{A_T}$ (\%)} \\
\midrule

CL Ours (Similar Tasks First)  & 1  & 60.75 \\
CL Ours (Reverse Order)        & 2  & 66.99 \\
CL Ours Avg                    & -- & 63.87 \\
\midrule
O-LoRA~\cite{wang-etal-2023-orthogonal} & 3 & 72.3 \\
O-LoRA~\cite{wang-etal-2023-orthogonal} & 4 & 64.8 \\
O-LoRA~\cite{wang-etal-2023-orthogonal} & 5 & 71.6 \\
O-LoRA Avg                     & -- & 69.6 \\
\midrule
\textbf{TORA}        & 1 & 75.45 \\
\textbf{TORA}        & 2 & 75.47 \\
\textbf{TORA}        & 3 & 75.23 \\
\textbf{TORA}        & 4 & 75.31 \\
\textbf{TORA}        & 5 & \textbf{75.56} \\
\textbf{TORA Avg}        & -- & \textbf{75.40} \\

\bottomrule
\end{tabular}


\caption{Detailed evaluation sequences for the static routing configurations.}
\label{tab:sequence_details}
\begin{tabular}{@{}ll p{0.80\textwidth}@{}}
\toprule
\textbf{Framework} & \textbf{Order} & \textbf{Task Sequence} \\
\midrule
\textbf{Ours} 
& \textbf{1} & Yelp $\to$ Amazon $\to$ IMDB $\to$ SST-2 $\to$ Yahoo $\to$ DBPedia $\to$ AGNews $\to$ QQP $\to$ BoolQ $\to$ MultiRC $\to$ WiC $\to$ RTE $\to$ COPA $\to$ MNLI $\to$ CB \\
& \textbf{2} & Exact reverse of Order 1 \\
\midrule
\textbf{O-LoRA} 
& \textbf{3} & MNLI $\to$ CB $\to$ WiC $\to$ COPA $\to$ QQP $\to$ BoolQ $\to$ RTE $\to$ IMDB $\to$ Yelp $\to$ Amazon $\to$ SST-2 $\to$ DBPedia $\to$ AGNews $\to$ MultiRC $\to$ Yahoo \\
& \textbf{4} & MultiRC $\to$ BoolQ $\to$ WiC $\to$ MNLI $\to$ CB $\to$ COPA $\to$ QQP $\to$ RTE $\to$ IMDB $\to$ SST-2 $\to$ DBPedia $\to$ AGNews $\to$ Yelp $\to$ Amazon $\to$ Yahoo \\
& \textbf{5} & Yelp $\to$ Amazon $\to$ MNLI $\to$ CB $\to$ COPA $\to$ QQP $\to$ RTE $\to$ IMDB $\to$ SST-2 $\to$ DBPedia $\to$ AGNews $\to$ Yahoo $\to$ MultiRC $\to$ BoolQ $\to$ WiC \\
\bottomrule
\end{tabular}

\end{table}

\vspace{-1cm}

\subsection{Geometric Phase Transition and the Divergence Threshold}

The threshold $\bar{\theta}_3 = 84.5^\circ$ emerges from the intrinsic geometry of the DistilBERT adapter space~\cite{aghajanyan2020intrinsic}. In high-dimensional spaces, orthogonal vectors ($90^\circ$) indicate complete independence between subspaces, a property that O-LoRA explicitly exploits by projecting new task updates onto the orthogonal complement of previous ones~\cite{wang-etal-2023-orthogonal}. TORA 
operates on the same geometric principle, but uses the angular distance as a continuous compatibility measure rather than enforcing strict orthogonality. A cosine similarity of $84.5^\circ$ ($\approx 0.095$) marks the practical boundary where shared variance becomes too small to support positive transfer. Below this threshold, the gradient directions~\cite{ruder2016overview} of the incoming task align sufficiently with the prior expert to enable knowledge reuse. Above it, forcing parameter transfer acts as a perturbation that requires the optimizer to unlearn irrelevant features, degrading performance.

\subsection{Structural Dynamics: Hubs vs. Isolated Nodes}

The routing decisions reveal two clearly distinct task roles. Broad datasets like DBPedia act as knowledge hubs, learning general geometric patterns that transfer easily to related domains and accelerate convergence. In contrast, tasks with strict objective functions such 
as MNLI, CB, and SST-2 develop incompatible parameter structures that require isolation. Forcing knowledge reuse in these cases causes geometric interference, confirming that the \textit{Shielding} mechanism is essential to prevent negative transfer.

\subsection{Computational Efficiency and Limitations}

Beyond mitigating catastrophic forgetting, TORA reduces computational 
overhead by initializing LoRA modules with geometric aligned weights, 
allowing tasks to skip the initial exploratory phase of gradient descent 
and reach convergence faster, as shown by the epoch savings reported 
in Table~\ref{tab:positive_transfer}. However, the current system uses 
a fixed neighborhood size $k=3$ and a static divergence threshold, 
which may limit scalability in longer task sequences. As the number of 
experts grows, a static $k$ risks capturing overlapping or contradictory 
gradient directions. Future work should explore dynamic neighborhood 
selection, such as adaptive clustering or density-based routing, to 
maintain transfer quality as the expert pool scales to 50 or more 
sequential tasks.

\section{Conclusion}

This work presented TORA, a geometric routing framework for continual learning in text classification that decides whether to transfer or isolate knowledge based solely on the angular distance between task fingerprints in the SVD parameter space. Evaluated across 15 diverse benchmarks, TORA never induced a harmful routing decision: compatible tasks benefited from the \textit{Boosting} protocol, saving up to 10 training epochs while exceeding isolated baselines, and structurally distant tasks were protected by \textit{Shielding}, preventing catastrophic forgetting without any accuracy loss. With a single geometric threshold of $\bar{\theta}_3 = 84.5^\circ$ and no reliance on task identities or predefined sequences, TORA demonstrates that the intrinsic geometry of adapter weights is a sufficient and reliable signal for autonomous knowledge routing in sequential text classification systems.

%

%

\bibliographystyle{splncs04} %
\bibliography{name} %

\end{document}